\PassOptionsToPackage{table}{xcolor}
\documentclass[10pt,twocolumn,letterpaper]{article}
\usepackage[T1]{fontenc}

\usepackage{cvpr}              %

\usepackage{booktabs}
\usepackage{multirow}
\usepackage[table]{xcolor}
\usepackage{graphicx}

\definecolor{cvprblue}{rgb}{0.21,0.49,0.74}
\usepackage[pagebackref,breaklinks,colorlinks,allcolors=cvprblue]{hyperref}

\def\paperID{18232} %
\def\confName{CVPR}
\def\confYear{2027}

\title{From Restoration to Redrawing: Revitalizing Ancient Chinese Characters by SVG Code Generation and Editing}

\author{
Weiguang Zhang\textsuperscript{1,2}\quad
Linhua Dong\textsuperscript{1,2}\quad
Bin Dong\textsuperscript{3}\quad
Shanshan Jiang\textsuperscript{3}\\
Maizhen Ning\textsuperscript{4}\quad
Xiaoqiang Kang\textsuperscript{1,2}\quad
Kun Zhang\textsuperscript{1,2}\quad
Hao Xu\textsuperscript{1,2}\\
Xiaowei Huang\textsuperscript{1}\quad
Wei Wang\textsuperscript{2}\quad
Kaizhu Huang\textsuperscript{5}\quad
Qiufeng Wang\textsuperscript{2}\\[0.5em]
{\small \textsuperscript{1}University of Liverpool} {\small \textsuperscript{2}Xi'an Jiaotong-Liverpool University}\\
{\small \textsuperscript{3}Ricoh Software Research Center Beijing Co., Ltd.}\\
{\small \textsuperscript{4}Tencent\quad
\textsuperscript{5}Duke Kunshan University}
}

\begin{document}
\maketitle

\begin{abstract}
Hanjian manuscripts written on Han Dynasty bamboo and wooden slips are invaluable heritage of ancient Chinese writing but suffer from damaged ink and incomplete strokes.
Existing image-to-image restoration methods may alter observed strokes and offer limited control over stroke geometry.
Inspired by experts' manual redrawing practices, we address these issues by introducing \textbf{ReDraw}, a two-stage vision-language model(VLM) that reformulates character restoration as Scalable Vector Graphics(SVG) code generation and editing.
Stage one targets faithful vectorization of observed strokes geometry or conservative tracing of mildly degraded strokes.
Guided by skeleton chain-of-thought reasoning and reinforcement learning with structure-aware rewards, stage two predicts skeletons before contours to redraw severely incomplete strokes by appending paths while keeping existing SVG paths unchanged. To support training, we develop a multi-layer data synthesis pipeline that disentangles ink, material, and degradation to generate 1.2M samples, and derive up to 6,400 adaptation pairs from 320 annotated real glyphs.
Experiments demonstrate synthetic restoration performance and adaptability to real Hanjian glyphs. Code and demo will be available upon acceptance.
\end{abstract}

\begin{figure}[!t]
  \centering
   \includegraphics[width=0.99\linewidth]{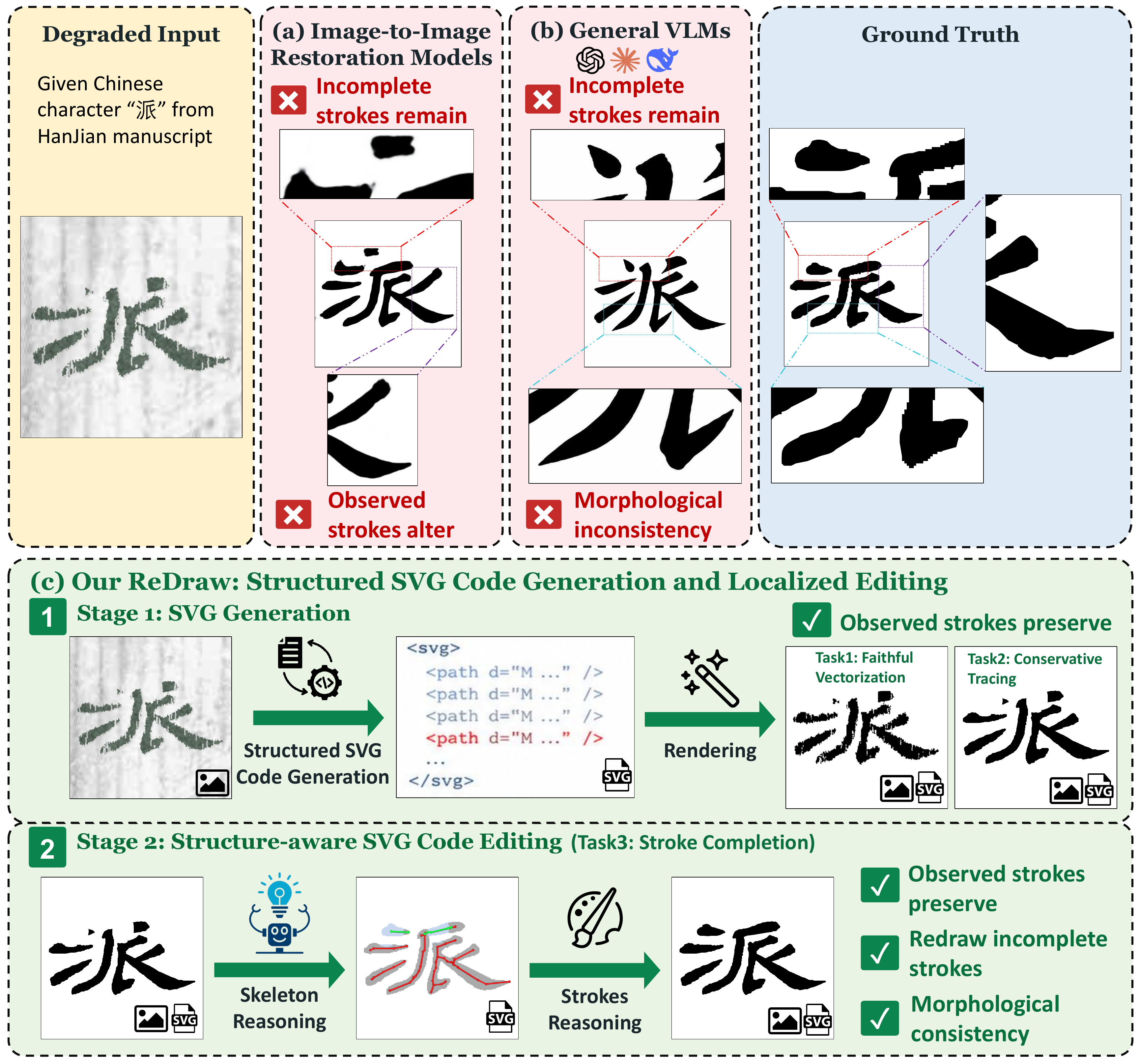}
   \caption{Comparison on ancient Chinese character restoration paradigms. Image-to-image restoration models (a) and general-purpose VLMs (b) may alter observed strokes or fail to restore severely incomplete strokes. ReDraw (c) generates SVGs through faithful vectorization or conservative tracing, then redraws incomplete strokes through skeleton-guided editing that appends paths while keeping existing SVG paths unchanged.
   }
   \label{fig:head}
\end{figure}    
\section{Introduction}
\label{sec:intro}
Chinese characters have been in use for over 3,000 years, and unearthed historical documents written in these characters serve as primary sources for understanding the history of ancient China~\cite{diao_oracle_2025}. 
Digitizing these unearthed documents serves as a prerequisite for downstream tasks, including deciphering ancient characters~\cite{liu2026alphaoracle}, tracing the evolution of Chinese writing systems~\cite{chi2022zinet}, analyzing calligraphic styles~\cite{liu2024deepcallifont}, and preserving and disseminating cultural heritage~\cite{li2026calliva}. 
However, centuries of physical and chemical deterioration have left these manuscripts with severe visual degradation, such as damaged ink, incomplete strokes, and damaged writing materials. Such degradation not only hinders the interpretation of ancient characters but also limits their reproduction for calligraphic study and publication. 
To address these challenges, we formulate ancient Chinese character restoration as redrawing, inspired by experts' practice of manually redrawing damaged strokes with digital editing software; our dataset focuses on Hanjian manuscripts~(Han Dynasty bamboo and wooden slips).

Existing approaches for ancient character visual restoration task have predominantly followed image-to-image translation paradigms~\cite{jiang2025survey}, leveraging document enhancement~\cite{jiang2027afg,ruan2026lwtformer,zhang_2024_docres}, inpainting~\cite{zhu2024text,wang2024coarse} and generative restoration~\cite{zhang2025bambink,yang2025diffhdr} to reconstruct single or full-page character images. 
Despite accomplishing notable advancements, their pixel-based formulation still presents some inherent issues. 
First, many of these approaches struggle to automatically identify regions requiring restoration while preserving observed strokes, potentially generating stylistically plausible characters with altered stroke structures, as shown in~\cref{fig:head}(a). Such holistic reconstruction may unnecessarily modify observed strokes, thereby compromising morphological consistency between the original and restored characters.
Second, character appearance in these approaches remains represented at the image level, limiting direct control over stroke geometry and calligraphic details. Although some methods~\cite{diao_oracle_2025,liu2025cinet} incorporate structural cues, they do not expose editable geometric primitives for localized completion of severely incomplete strokes.
These limitations motivate a fundamental reconsideration: \textit{Can we move beyond pixel-based restoration by representing ancient characters in a structured and editable form?}

 To this end, we propose \textbf{ReDraw}, the first effort that reformulates ancient character restoration as a Scalable Vector Graphics (SVG) code generation and editing task. 
 Unlike pixel-based representations, SVG encodes character shapes as structured geometric primitives and paths, enabling resolution-independent rendering and fine-grained structural manipulation. 
 Specifically, these geometric paths allow traced strokes and proposed completions to be represented as distinct, editable code segments, thereby shifting character restoration from holistic image reconstruction to localized code generation.
 Meanwhile, recent advances in vision-language models (VLMs) have demonstrated remarkable capabilities across diverse tasks, \eg code generation~\cite{ma2026repositories,wang2026internsvg,Sun_2026_CVPR} and visual reasoning~\cite{lu2026visualprimitives,wu2025spatialreasoning}. By integrating structured SVG representations into VLMs, ReDraw establishes a unified framework for diverse restoration tasks, leveraging structure-aware reasoning to reconstruct severely incomplete strokes.

 Building upon these insights, ReDraw adopts a two-stage SVG-based framework that unifies three subtasks for different restoration requirements(See~\cref{sec:data_preliminary}). 
 In the first stage, SVG code generation supports two subtasks: faithful vectorization, which aims to generate the observed strokes geometry and its degradation traces, and conservative tracing, which redraws those mildly damaged ink while leaving severely incomplete stroke for subsequent completion.
 In the second stage, SVG code editing redraws incomplete stroke segments by appending new paths while leaving the given SVG paths unchanged. To facilitate this process, we introduce a skeleton chain-of-thought (CoT)~\cite{wei2022cot} reasoning strategy that predicts skeletons of incomplete stroke segments before generating their contours, and further enhance structural completion through reinforcement learning~\cite{shao2024deepseekmath} with structure-aware rewards. 
 To support these subtasks, we develop a multi-layer data synthesis pipeline that disentangles ink, material, and degradation into three independent layers, allowing flexible and realistic data composition. Through this pipeline, we generate 1.2 million synthetic samples and employ multi-task fine-tuning to train a unified VLM for SVG generation and editing. Additionally, we curate an expert-verified real-world Hanjian dataset, comprising 320 real glyphs, from which 6,400 training pairs are derived for few-shot glyph adaptation.

In summary, our main contributions are four-fold:
\begin{itemize}[topsep=1pt,leftmargin=*]
\item We introduce ancient Chinese character redrawing, the first paradigm that reformulates existing pixel-based restoration task as structured SVG generation and editing.

\item We construct a Hanjian dataset comprising 1.2 million synthetic samples generated from a multi-layer synthesis pipeline, alongside 320 annotated real-world glyphs yielding up to 6,400 adaptation pairs.

\item We develop a unified two-stage VLM framework for SVG generation and editing, enhanced by skeleton chain-of-thought reasoning and reinforcement learning for character structural completion.

\item Extensive experiments demonstrate the effectiveness and adaptability of our framework across synthetic and real-world Hanjian manuscript.
\end{itemize}

\section{Related Work}
\label{sec:relat}
\subsection{Historical Document Restoration}
Historical document restoration primarily follows two sequential steps, \ie, missing textual content restoration followed by visual appearance restoration. Existing methods address either step separately or integrate both.

\textbf{Textual content restoration.}
Pythia~\cite{assael2019pythia} pioneers textual content restoration by inferring missing Greek characters from linguistic context.
Subsequent work advanced this direction through adaptation to Babylonian and Latin texts~\cite{fetaya2020babylonian,bamman2020latinbert}, multilingual pretraining for Akkadian gap filling~\cite{lazar2021akkadian}, and joint geographical and chronological attribution~\cite{assael2022ithaca}.
Our work instead focuses on second-step visual appearance restoration given known character identities.

\begin{figure*}[!ht]
	\centering
	\includegraphics[width=\textwidth]{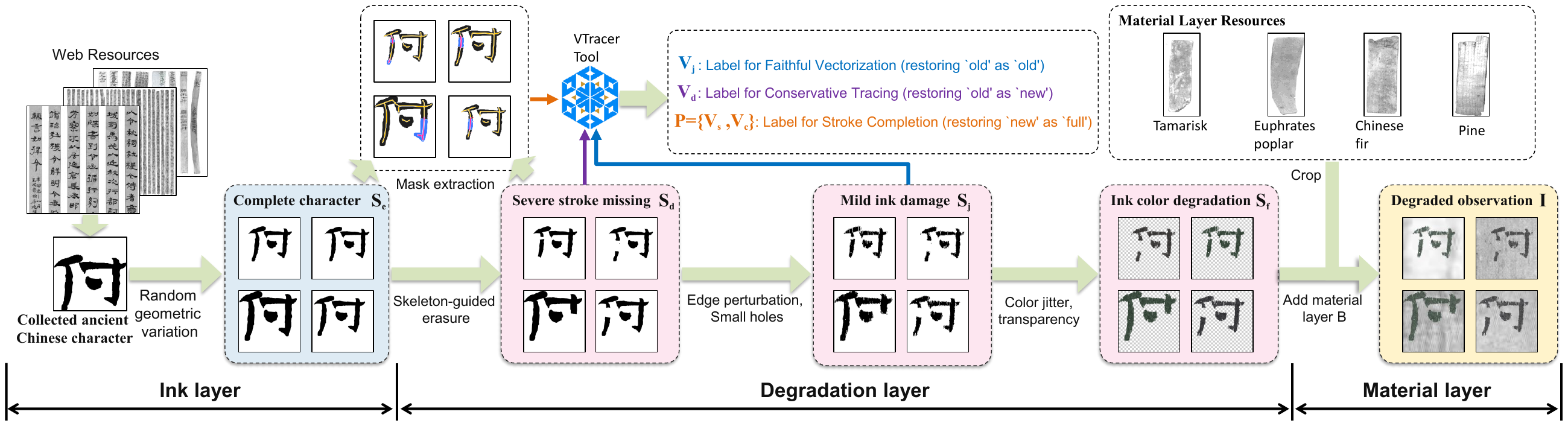}
	\caption{\textbf{The pipeline of multi-layer data synthesis.} Ink, degradation, and material layers produce degraded observations with paired SVG labels for three subtasks: faithful vectorization, conservative tracing, and skeleton-guided stroke completion.}
	\label{fig:dataset_pipeline}
\end{figure*}

\textbf{Visual appearance restoration.}
Following advances in Transformer-based natural image restoration, exemplified by SwinIR~\cite{liang_2021_swinir} and Uformer~\cite{wang_2022_uformer}, one line such as CharFormer~\cite{shi_2022_charformer} and DocRes~\cite{zhang_2024_docres} adopt Transformer backbones for character and full-page restoration, respectively.
Another line explores generative visual appearance restoration, including GAN-based ink enhancement in BambInk~\cite{zhang2025bambink} and diffusion-based restoration in DocDiff~\cite{yang_2023_docdiff}, DiffACR~\cite{li2024diffacr}, and DiffHDR~\cite{yang2025diffhdr}.
Recent studies further emphasize structure-aware character restoration. For instance, AFG-Net~\cite{jiang2027afg} improves glyph preservation through frequency-domain hard-sample learning, while LWTformer~\cite{ruan2026lwtformer} preserves stroke details through learnable wavelet modeling.
For explicit glyph guidance, Diao \etal~\cite{diao_oracle_2025} restore oracle bone inscriptions through joint glyph prediction, while CINet~\cite{liu2025cinet} uses ideographic description sequences (IDS) as structural representations to guide inpainting.
However, IDS captures general character composition without distinguishing fine-grained strokes details, while raster glyphs don't provide editable strokes.
These methods still restore raster images; to our knowledge, SVG generation and editing remain unexplored for visual appearance restoration of ancient Chinese characters.

\textbf{Joint restoration.}
Joint methods integrate textual and visual restoration.
Zhu \etal~\cite{zhu2024reproducing} adopt diffusion with a retrieval-guided text correction, while AutoHDR~\cite{zhang2025autohdr} couples LLM predictions with patch-autoregressive visual appearance restoration.
More recently, Zhu \etal further introduce EpiAgent~\cite{zhu2026epiagent}, an agentic framework for ancient inscription restoration through adaptive tool coordination.
UniHIR~\cite{zhang2026unihir} develops a unified multimodal understanding and generation model for joint restoration.
Despite promising results from multimodal model and agentic coordination, these methods restore appearance in pixel space, whereas ReDraw uses VLMs to generate and edit SVG paths for localized stroke restoration.

\subsection{Image-to-Code Generation}
\textbf{Complex Document Element Representation.}
Image-to-Code generation represents complex document element in structured forms that support better perception, understanding and reasoning.
Recent work explores diverse code representations, including TikZ for geometric diagrams~\cite{Sun_2026_CVPR}, Python for charts~\cite{zhao2025chartcoder}, HTML for tables~\cite{peng2024unitable}, atom--bond graphs for molecular diagrams~\cite{qian2023molscribe}, and parametric primitives for CAD analysis~\cite{niu2025phtcad,qi2026pointercad}.
For character representation, DeepVecFont series~\cite{wang2021deepvecfont,wang2023deepvecfontv2} combine raster appearance and vector sequence modeling to synthesize SVG.
OracleFusion~\cite{li2025oraclefusion} combines multimodal reasoning with structurally controlled SVG generation to assist oracle bone decipherment.
Similarly, ReDraw extends structured representations to historical document restoration, using SVG code for explicit stroke geometry and localized editing.

\textbf{Image Vectorization.}
Image vectorization broadly follows optimization or LLM-based modeling.
DiffVG~\cite{li2020diffvg} and LIVE~\cite{ma2022live} optimize vector paths through differentiable rasterization, while VectorFusion~\cite{jain2023vectorfusion} incorporates diffusion priors for text-conditioned synthesis.
LLM-based methods generate SVG by VLMs, including StarVector~\cite{rodriguez2025starvector}, OmniSVG~\cite{yang2025omnisvg} and InternSVG~\cite{wang2026internsvg}. 
CTRL-S~\cite{wang2026ctrls} and RefineSVG~\cite{liu2026refinesvg} further improve generation with reinforcement learning.
For characters, Vector Calligrapher~\cite{zhou2026vectorcalligrapher} studies clean Chinese character vectorization, while LVGM~\cite{zhang2025lvgm} encodes  complete stroke as learned codebook for stroke-sequence continuation, without addressing localized stroke damage. Unlike these works, our ReDraw instead uses image vectorization to repair damaged strokes in real-world ancient Chinese character.

\begin{figure*}[!ht]
	\centering
	\includegraphics[width=\textwidth]{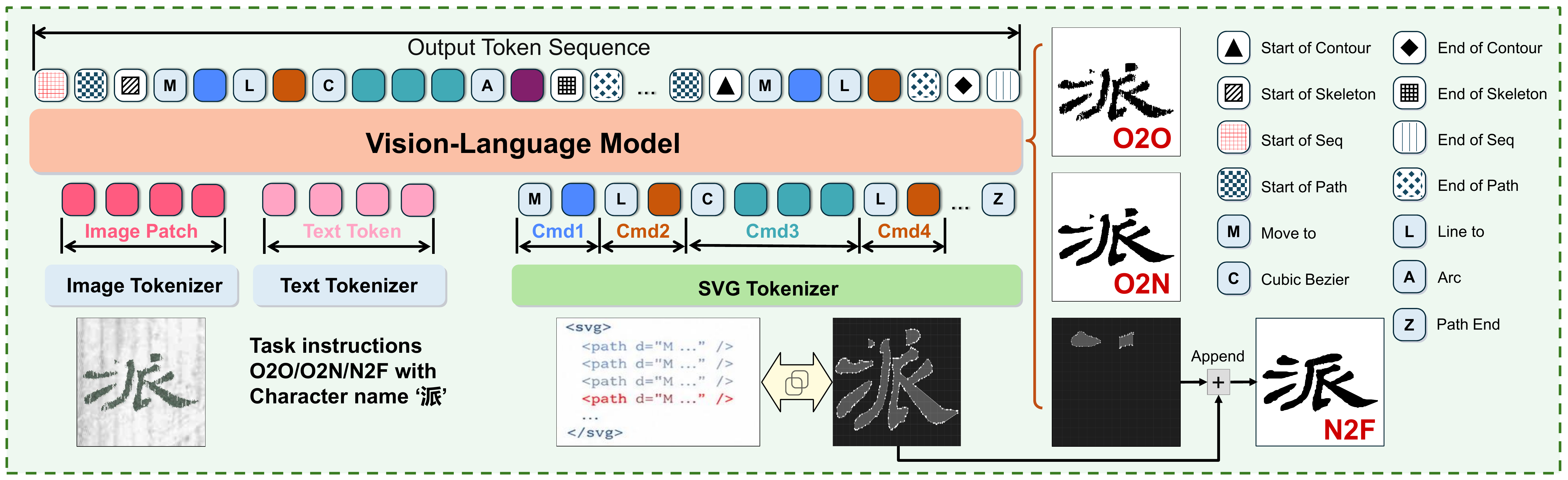}
	\caption{\textbf{Overview of ReDraw.} A shared VLM generates SVGs for faithful vectorization or conservative tracing first, and then predicts interleaved skeleton--contour paths for stroke completion, appending only contours while preserving existing paths.}
	\label{fig:method}
\end{figure*}

\section{Methodology}
\label{sec:data}
\subsection{Task Definition}
\label{sec:data_preliminary}
Due to the fragility of unearthed materials, historical document conservation has long valued preserving its original state, often expressed as \emph{restoring `old' as `old'}, together with the principle of minimal intervention~\cite{du1999restoration,wang2021minimal,ICOMOS_1964_Venice}. 
Such principle reduces the risk of losing historical evidence through repeated physical treatment, while limiting the possibility of making full use of ancient character. Digital restoration offers a more flexible way to revitalize ancient characters: candidate reconstructions can be generated, compared, and revised without altering the physical originals.
To accommodate these restoration requirements, we define three subtasks: faithful vectorization, conservative tracing, and stroke completion.
The first two constitute stage one, which generates SVG representations from degraded images; the third constitutes stage two, which completes damaged strokes through SVG editing with known class identities.

In stage one, an ancient Chinese character image $I$ is converted into an SVG under either of two subtasks. \textbf{Faithful vectorization}, termed \emph{restoring `old' as `old'} (\textbf{O2O}), maps $I$ to $V_j$, which extracts the observed strokes geometry while retaining its degradation traces, including worn contours, small holes, and broken stroke \etal. 
\textbf{Conservative tracing}, termed \emph{restoring `old' as `new'} (\textbf{O2N}), maps the same $I$ to $V_d$, which additionally repairs mild geometric degradation, such as local contour erosion and small holes, while leaving severely incomplete strokes unresolved. Thus, `new' denotes a cleaned but damaged character with incomplete glyph, and the two objectives differ in the geometric degradation retained in their targets. 
Stage two performs \textbf{stroke completion}, termed \emph{restoring `new' as `full'} (\textbf{N2F}).
Given an incomplete SVG $V_d$ from O2N and its raster image $S_d$, this subtask predicts a code patch $P= \{V_s,V_c\}$ containing additional paths for missing stroke segments, denoted by $(V_d,S_d) \mapsto P$.
Skeleton predictions $V_s$ guide the generation of their contours $V_c$.
Here, `full' denotes structural completion; the added paths remain editable candidate reconstructions, not verified original historical appearance.

\begin{table*}[tp]
	\centering
	\caption{Comparison of current vectorization models and our ReDraw on 600 synthetic test samples. Time and token counts are measured only for O2O. Best and second-best performance are marked in bold and underlined, respectively.}
	\label{tab:comparison1}
	
	\resizebox{\textwidth}{!}{%
		\begin{tabular}{l|cc|ccc|ccc|ccc}
			\toprule
			\multirow{2}{*}{Method}
			& \multicolumn{2}{c|}{Avg. Efficiency}
			& \multicolumn{3}{c|}{Restoring 'old' as 'old'}
			& \multicolumn{3}{c|}{Restoring 'old' as 'new'}
			& \multicolumn{3}{c}{Restoring 'new' as 'full'} \\
			\cmidrule(lr){2-3}
			\cmidrule(lr){4-6}
			\cmidrule(lr){7-9}
			\cmidrule(lr){10-12}
			& Time/s & Tokens
			& SSIM$\uparrow$ & LPIPS$\downarrow$ & MSE$\downarrow$
			& SSIM$\uparrow$ & LPIPS$\downarrow$ & MSE$\downarrow$
			& SSIM$\uparrow$ & LPIPS$\downarrow$ & MSE$\downarrow$ \\
			\midrule
			
			\rowcolor{gray!25}
			\multicolumn{12}{c}{\textit{General-purpose  VLMs}} \\
			\midrule
			
			GPT-6 Astra~\cite{openai2026gpt6astra}
			& 50.3 & 2169
			& \underline{0.8546} & \underline{0.1768} & \underline{0.0334}
			& \underline{0.8814} & \underline{0.1072} & \textbf{0.0247}
			& \underline{0.8983} & \underline{0.1396} & \underline{0.0456} \\
			
			Claude Opus 5.5~\cite{anthropic2026opus55}
			& 16.6 & 1305
			& 0.6737 & 0.5651 & 0.1482
			& 0.8155 & 0.1823 & 0.0558
			& 0.8413 & 0.2125 & 0.0751\\
			
			DeepSeek-V4.1-Flash~\cite{deepseek2026v41flash}
			& 34.0 & 3643
			& 0.7241 & 0.3697 & 0.1059 
			& 0.7260 & 0.3248 & 0.1151
			& 0.8958 & 0.1677 & 0.0579 \\
			
			Kimi K3~\cite{moonshot2026kimik3}
			& 58.1 & 1054
			& 0.7256 & 0.3985 & 0.1070
			& 0.7036 & 0.3398 & 0.1288
			& 0.8897 & 0.1671 & 0.0565\\
			
			GLM-5.3-Flash~\cite{zai2026glm53flash}
			& 7.5 & 631
			& 0.7167 & 0.4225 & 0.1135
			& 0.7168 & 0.3581 & 0.1209
			& 0.8610 & 0.1869 & 0.0654 \\
			
			\midrule
			\rowcolor{gray!25}
			\multicolumn{12}{c}{\textit{Existing Image-to-SVG Models}} \\
			\midrule
			DiffVG~\cite{li2020diffvg}
			& 84.7 & 15949
			& 0.6230 & 0.3958 & 0.0920
			& --- & --- & ---
			& --- & --- & --- \\

			OmniSVG~\cite{yang2025omnisvg}
			& 164.0 & 5228
			& 0.4519 & 0.7484 & 0.2799
			& --- & --- & ---
			& --- & --- & --- \\

			CTRL-S~\cite{wang2026ctrls}
			&272.8 & 7237
			&0.6852 &0.5865 &0.1362
			& --- & --- & ---
			& --- & --- & --- \\

			RefineSVG~\cite{liu2026refinesvg}
			& 64.4 & 2909
			& 0.5792 & 0.6616 & 0.1283
			& --- & --- & ---
			& --- & --- & --- \\

			\textbf{Our model:~ReDraw}
			& 98.8 & 4813
			& \textbf{0.9375} & \textbf{0.0609} & \textbf{0.0145}
			& \textbf{0.9196} & \textbf{0.0869} & \underline{0.0259}
			& \textbf{0.9860} & \textbf{0.0374} & \textbf{0.0110} \\

			\bottomrule
		\end{tabular}%
	}
\end{table*}

\subsection{Multi-layer Data Synthesis Pipeline}
\label{sec:data_synthesis}
Annotating real ancient character images requires labor-intensive manual tracing with image editing tools to recover clean characters from degraded observations.
Inspired by this practice, we reverse the process, automatically composing ink, material, and degradation layers from clean characters to approximate the appearance of real manuscripts while retaining paired supervision (See~\cref{fig:dataset_pipeline}).

\textbf{Ink layer.}
To capture the variation associated with \emph{libian}, the evolution from seal to clerical forms reflected in Hanjian writing~\cite{zhao2020libian,li2025dunhuang}, we collect 120k seal and clerical script vector character from web resources.
These characters cover 6,924 character classes, with around 17 glyph variants per class.
For each glyph, we generate ten variants through random geometric and stroke-width variation, yielding 1.2M clean raster references $S_e$.

\textbf{Material layer.}
We collect 400 character-free slip images from the DeepJiandu dataset~\cite{liu2025deepjiandu} and group them into four material categories: pine, Euphrates poplar, Chinese fir, and tamarisk.
We randomly select a category and crop the texture to match the ink-layer dimensions, and use it to fill the transparent background layer $B$.

\textbf{Degradation layer.}
We simulate two levels of stroke damage followed by color and opacity degradation.
For severe damage, we randomly erase local stroke segments along their skeletons or remove all ink within a longitudinal strip to simulate damage along the wood grain, producing incomplete glyphs $S_d$ from $S_e$.
For mild damage, we further apply edge perturbation and small-hole corruption to $S_d$, yielding $S_j$.
Subsequent color and opacity variations produce ink layers $S_f$, which are composited with $B$ into observations $I$ at $448\times448$ resolution.
Since these operations take place in pixel space, we use VTracer\footnote{\url{https://github.com/visioncortex/vtracer}} to vectorize $S_d$ and $S_j$ into $V_d$ and $V_j$.
Each $I$ is paired with $V_j$ for O2O and $V_d$ for O2N.
For N2F, the difference between $S_e$ and $S_d$ gives a mask for missing stroke, which we vectorize with VTracer to obtain contour paths $V_c$.
We apply Zhang--Suen thinning~\cite{zhang1984thinning} to this mask and encode the resulting skeletons as paths $V_s$, forming the patch $P=\{V_s,V_c\}$.

\subsection{Model Construction}
\label{sec:method}
\textbf{Model architecture.}
ReDraw builds on Qwen2.5-VL-3B~\cite{bai2025qwen25vl}, combining its visual encoder and autoregressive language decoder with an extended SVG tokenizer. As shwon in~\cref{fig:method}, a SVG tokenizer serializes SVG paths into command and coordinate tokens, whose embeddings map SVG geometry into the VLM's unified representation space.
The $j$-th command in the $i$-th path is represented as $C_{ij}=(U_{ij},V_{ij})$, where $U_{ij}\in\{M,L,C,A,Z\}$ denotes its type and $V_{ij}$ its geometric arguments.
Following IconShop~\cite{wu2023iconshop} and OmniSVG~\cite{yang2025omnisvg}, we encode each coordinate pair as a single token, using a $200\times200$ quantization grid in our implementation.
For O2O and O2N, the input prefix combines a task instruction with $I$ to generate $V_j$ and $V_d$, respectively.
For N2F, the prefix combines a task instruction with character class, the raster image $S_d$, and tokens of the incomplete SVG $V_d$.
Special delimiters serialize the predicted patch $P$ into interleaved skeleton--contour pairs $(V_s^{(1)},V_c^{(1)},\ldots,V_s^{(K)},V_c^{(K)})$, where $K$ is the number of completion patches.
Each contour is conditioned on its preceding skeleton and previously generated patches, making skeleton CoT an explicit geometric intermediate.
At inference, N2F receives the predicted O2N SVG and its rasterization from stage one. Post-processing appends only contour paths, preserving the input SVG paths at the code level. Geometric fidelity still depends on first-stage tracing errors and the placement of added paths.

\begin{figure*}[!t]
	\centering
	\includegraphics[width=0.94\textwidth]{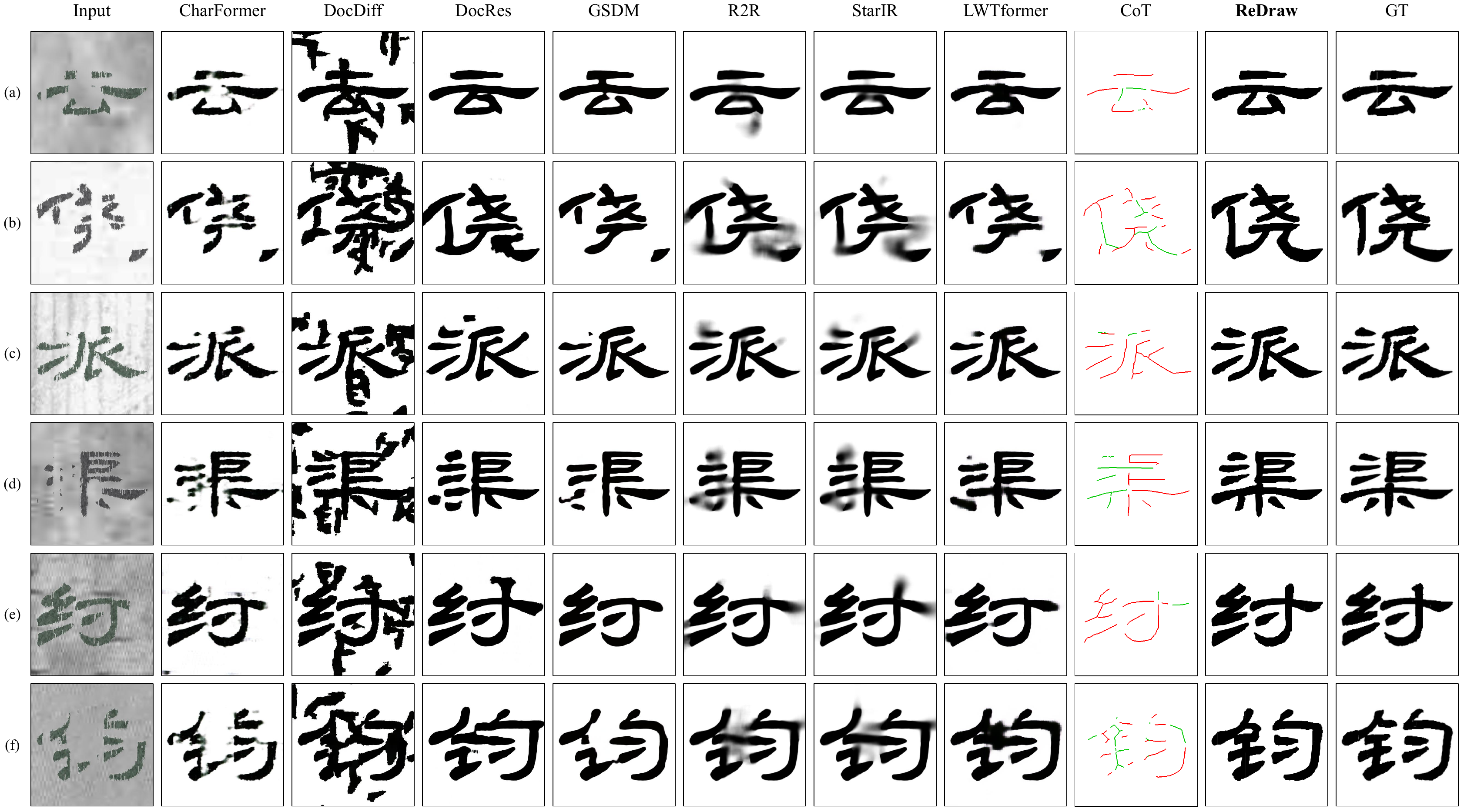}
	\caption{\textbf{Qualitative comparison on synthetic character restoration.} These examples compare image-to-image baselines with ReDraw's completed results and clean references (GT). ReDraw restore incompleted stroke segments while retaining observed strokes.}
	\label{fig:comparison}
\end{figure*}

\textbf{Training objectives.}
Supervised fine-tuning (SFT) learns vectorization and skeleton-guided completion through the same next-token objective:
\begin{equation}
\mathcal L_{\mathrm{SFT}}=-\mathbb E_{(x,y)}\frac{1}{T}
\sum_{t=1}^{T}\log\pi_\theta(y_t\mid x,y_{<t}),
\label{eq:sft}
\end{equation}
where $x$ is the task-specific input and $y$ the target SVG sequence of length $T$.
Only output tokens, including skeleton and contour delimiters, contribute to the loss.

To align generated geometry beyond token matching, we further refine N2F with group relative policy optimization (GRPO)~\cite{shao2024deepseekmath}.
Rewards use masks rendered at $256\times256$.
Skeleton reward $r_s$ compares dilated line masks with the target skeleton; contour reward $r_c$ compares filled masks with the target contour, both using IoU.
The consistency reward compares the predicted skeleton with the skeleton extracted from its predicted contour:
\begin{equation}
r_{sc}=\operatorname{IoU}\!\left(
\mathcal D[\mathcal R(\widehat V_s)],
\mathcal D[\mathcal S(\mathcal R(\widehat V_c))]\right),
\label{eq:skeleton_consistency}
\end{equation}
where \(\mathcal R\) renders skeleton paths as thin strokes and contour paths as filled regions, $\mathcal S$ extracts skeleton from contour mask, and $\mathcal D$ dilates it.
The merge reward $r_m$ measures the IoU between the masks obtained by adding the predicted and target patches, respectively.
These rewards aggregate paths across completion segments.
Invalid outputs receive $-1$; valid outputs receive format reward $r_f=1$ and total reward is denoted as:
\begin{equation}
R=\operatorname{clip}_{[-1,1]}\!\left[
g_s\!\left(\mathbf w^\top\mathbf r-\lambda_o p_o-\lambda_a p_a\right)\right],
\label{eq:structure_reward}
\end{equation}
where $\mathbf r=(r_f,r_s,r_c,r_{sc},r_m)$.
We set $\mathbf w=(0.1,0.3,0.2,0.3,0.1)$, $\lambda_o=0.1$, and $\lambda_a=0.05$.
Here $p_o$ is the IoU between the predicted patch and existing ink, and $p_a=\min(1,|\log((\widehat a+\epsilon)/(a+\epsilon))|)$ penalizes patch-area mismatch, with $\widehat a$ and $a$ denoting predicted and target areas.
The gate $g_s$ is $0.1$ when the predicted skeleton is empty despite a nonempty target, and $1$ otherwise.
For each input, GRPO samples a group of COT complete sequences and computes advantages by normalizing their rewards within the group.
These advantages guide a single policy update without probability-ratio clipping, with a KL penalty limiting deviation from the fixed SFT reference model.

\begin{table}[!t]
	\centering
	\caption{Comparison with Image-to-image restoration on the synthetic test set. All methods use the same degraded inputs and clean, complete character references. Best and second-best performance are marked in bold and underlined, respectively.}
	\label{tab:img2img_comparison}
	\resizebox{\columnwidth}{!}{
		\begin{tabular}{llrrr}
			\toprule
			Method & Venue & SSIM$\uparrow$ & LPIPS$\downarrow$ & MSE$\downarrow$ \\
			\midrule
			SwinIR~\cite{liang_2021_swinir} & ICCVW'21 & 0.8458 & 0.1747 & 0.04781 \\
			Uformer-B~\cite{wang_2022_uformer} & CVPR'22 & 0.8375 & 0.1438 & 0.03580 \\
			CharFormer~\cite{shi_2022_charformer}& ACM MM'22 & 0.8033 & 0.2145 & 0.05164 \\
			PromptIR~\cite{potlapalli_2023_promptir} & NeurIPS'23 & 0.8707& \underline{0.1321} & 0.03224 \\
			DocDiff~\cite{yang_2023_docdiff} & ACM MM'23 & 0.6260 & 0.3771 & 0.17152 \\
			DocRes~\cite{zhang_2024_docres} & CVPR'24 & \underline{0.8732} & 0.1367 & 0.02813 \\
			GSDM~\cite{zhu2024text} & AAAI'24 & 0.8581  &0.1472 &0.03435 \\
			AFM~\cite{ryou_2024_afm} & CVPR'24 & 0.5850 & 0.3101 & 0.04675 \\
			MambaIRv2~\cite{guo_2025_mambairv2} & CVPR'25 & 0.8468 & 0.1694 & 0.04683 \\
			AdaIR~\cite{cui_2025_adair} & ICLR'25 & 0.8664 & 0.1380 & 0.03329 \\
			SFIR~\cite{gu_2026_sfir} & PR'26 & 0.8250 & 0.2046 & 0.02795 \\
			R2R~\cite{wang_2026_r2r} & CVPR'26 & 0.8459 & 0.1863 & \underline{0.02498}\\
			StarIR~\cite{cui_2026_starir} & TPAMI'26 & 0.8706 &0.1647 & \textbf{0.01960} \\
			LWTformer~\cite{ruan2026lwtformer} & CVPR'26 &0.8575  &0.1389  &0.03241  \\
			\midrule
			\textbf{Our ReDraw} & - & \textbf{0.9225}  & \textbf{0.0850}  & 0.02832 \\
			\bottomrule
	\end{tabular}}

\end{table}

\section{Experiments}
\label{sec:experiments}
To validate the effectiveness of our method, we first evaluate ReDraw on synthetic character redrawing across O2O, O2N, and N2F, compare it with image-to-image restoration methods, and then examine few-shot adaptation to real Hanjian glyphs.
We rasterize generated SVGs and compare them with task-specific references using SSIM~\cite{wang2004ssim} for structural similarity, LPIPS~\cite{zhang2018lpips} for perceptual fidelity, and MSE for pixel-wise error.
Higher SSIM and lower LPIPS/MSE indicate better results.
For N2F, we evaluate the completed glyph after appending predicted contours to the incomplete SVG.

\subsection{Implementation Details}
\label{sec:implement}
We train ReDraw on two NVIDIA A800 GPUs in bfloat16 precision with four-step gradient accumulation.
We adopt the AdamW optimizer with an initial learning rate of $1\times10^{-4}$, cosine learning-rate decay, and a weight decay of $0.1$ for both supervised stages, training each for three epochs.
In practice, we initialize ReDraw from Qwen2.5-VL-3B and initialize the newly added SVG embeddings from scratch, with input images resized to $448\times448$.
For the main results, GRPO reuses the 1.2M-sample SFT data pool for 10,000 additional updates at a learning rate of $1\times10^{-5}$; the 60K-data ablation uses 1,000 additional GRPO updates.
Unless otherwise specified, we generate SVGs using top-$k$ and top-$p$ sampling with $k=50$, $p=0.90$, a temperature of $0.3$, and a repetition penalty of $1.05$. More experimental details can be obtained in the supplementary materials.

\begin{figure}[!t]
	\centering
	\includegraphics[width=\linewidth]{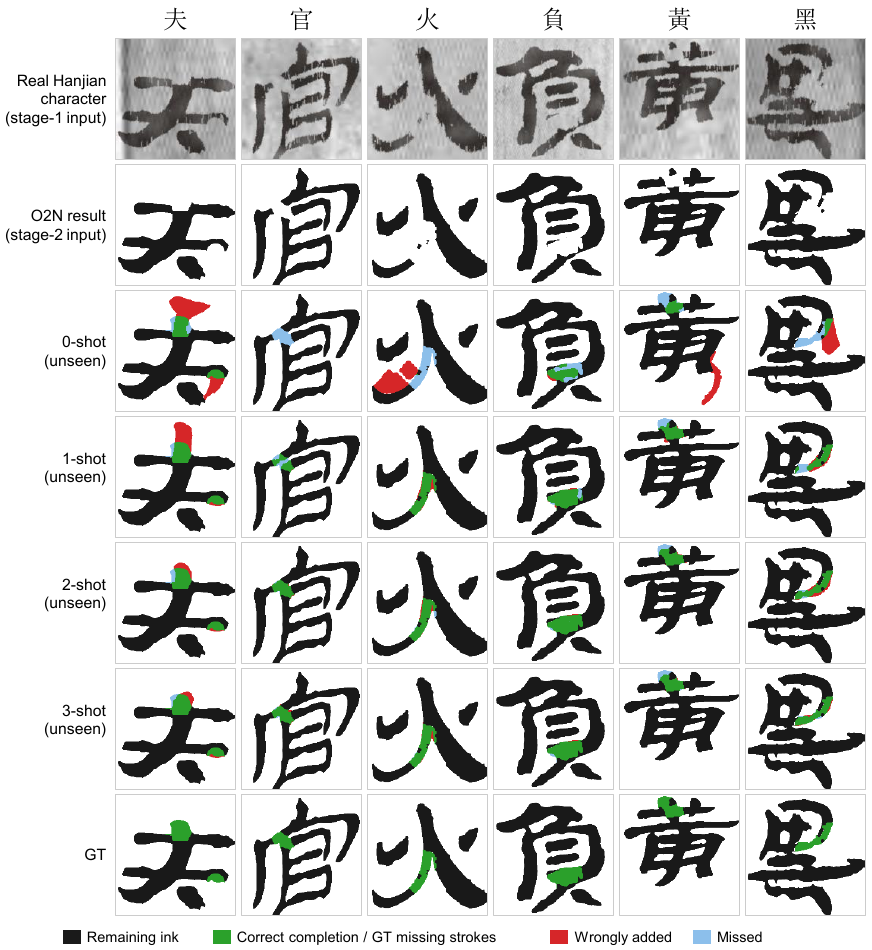}
	\caption{\textbf{Qualitative results of synthetic-to-real adaptation on unseen character glyphs.} Black denotes observed stroke ink, green represents correctly completed regions (missing strokes in GT), red represents incorrect additions, and blue shows missed regions. 
	As the number of shots increases, the green areas gradually expand, while the red and blue areas gradually decrease.}
	\label{fig:fewshot}
\end{figure}

\subsection{Main Results}
\subsubsection{Synthetic Domain Evaluation}
\textbf{Comparison with image vectorization models.}
We evaluate on 600 synthetic samples sharing source glyphs with training but differing in damage pattern and backgrounds, testing general restoration capacity rather than unseen glyphs. In \cref{tab:comparison1}, baselines receive no task-specific training.
General-purpose VLMs are evaluated on all three tasks; specialized image-to-SVG methods use their default settings and vectorization instructions for O2O only.
Inference time and token counts are measured only for O2O; additional visual comparisons appear in Supplementary Material~1.  Among general-purpose VLMs, GPT-6 Astra stands out from the other evaluated models, particularly on O2O: its LPIPS is 0.1768, compared with 0.3697 for the second-best general-purpose model.
The specialized SVG models even underperform GPT-6 Astra on O2O, indicating that existing image-to-SVG model does not ensure generalization to ancient character images or degraded image background.
ReDraw, well trained on the same data distribution with the test set, leads on all quality metrics except O2N MSE (0.0259 versus GPT-6 Astra's 0.0247), supporting its effectiveness within this synthetic setting.

\begin{table}[!t]
	\centering
	\caption{Synthetic-to-real adaptation: Seen/unseen refer to glyphs within the same characters; their subsets vary with $k$. The 0-shot row covers all 80 character classes(correspond to 320 real glyphs). }
	\label{tab:fewshot_adaptation}
	\footnotesize
	\setlength{\tabcolsep}{5pt}
	\begin{tabular}{@{}lrrrrrr@{}}
		\toprule
		\multirow{2}{*}{Setting} & \multicolumn{3}{c}{Seen glyphs} & \multicolumn{3}{c}{Unseen glyphs} \\
		\cmidrule(lr){2-4}\cmidrule(lr){5-7}
		& SSIM$\uparrow$ & LPIPS$\downarrow$ & MSE$\downarrow$ & SSIM$\uparrow$ & LPIPS$\downarrow$ & MSE$\downarrow$ \\
		\midrule
		0-shot & --- & --- & --- & 0.9508 & 0.0642 & 0.02387 \\
		1-shot & 0.9769 & 0.0330 & 0.01032 & 0.9721 & 0.0389 & 0.01042 \\
		2-shot & 0.9779 & 0.0349 & 0.00802 & 0.9767 & 0.0311 & 0.00933 \\
		3-shot & 0.9810 & 0.0283 & 0.00664 & 0.9760 & 0.0274 & 0.00783 \\
		4-shot & 0.9812 & 0.0261 & 0.00717 & --- & --- & --- \\
		\bottomrule
	\end{tabular}
\end{table}

\textbf{Comparison with image-to-image restoration.}
We compare ReDraw with open-source document and general image restoration networks in~\cref{tab:img2img_comparison,fig:comparison}.
For fair comparison, we re-train these models and ReDraw on the same 60k synthetic samples for 10,000 steps.
All methods receive the same degraded observations and are evaluated against clean, complete character references.
ReDraw achieves the best SSIM (0.9225) and LPIPS (0.0850), improving SSIM by 0.0493 over DocRes and reducing LPIPS by 35.7\% relative to PromptIR. \cref{fig:comparison} provides a more intuitive comparison. We selected several models with better performance from~\cref{tab:img2img_comparison}. Our ReDraw produces more complete strokes while better retaining existing geometry than baselines exhibiting missed completions, blur, or altered strokes.
Local shape and connection errors remain in rows (b), (d), and (f), potentially contributing to its higher MSE than StarIR (0.02832 versus 0.01960).

\subsubsection{Synthetic-to-Real Adaptation}
\label{sec:fewshot_adaptation}
We further investigate whether our ReDraw model with synthetic data can adapt to real ancient Chinese character and generalize across different glyphs of the same character class.
These adaptation focus on N2F task. At inference time, its partial SVG and raster input are obtained from predicted O2N outputs from stage one. 
We collect 80 character classes with four real glyphs each, totaling 320 glyphs. 
We apply super-resolution to the collected images as a preprocessing step before manual completion.
Then, we employ four volunteers and spent around 200 working hours manually completing these images and seeking experts to review.
For $k$-shot adaptation, we select $k$ annotated glyphs per character and generate 20 training variants per glyph, yielding $1{,}600k$ pairs (up to 6,400).
Thus, each shot denotes a distinct glyph, rather than an augmented training pair. Seen glyphs share their source glyph with adaptation data but have independently sampled damage; unseen glyphs are excluded from adaptation data while belonging to the same character class.
As shown in~\cref{tab:fewshot_adaptation}, without adaptation, the pretrained ReDraw model underperforms for real character.
Even after one-shot adaptation, Unseen glyphs achieve SSIM/LPIPS/MSE of 0.9721/0.0389/0.01042, suggesting that adaptation benefits can extend beyond the training glyphs.
The qualitative comparisons in~\cref{fig:fewshot} complement these results: adaptation reduces incorrect additions and missed stroke regions on the illustrated unseen glyphs, while some local completion errors persist.
These results support inter-class glyph adaptation; changing test subsets across shot counts prevents a controlled comparison of the effect of additional shots.

\subsection{Ablation Study}
\label{sec:ablation}
\textbf{Effectiveness of stage decomposition.}
Our two-stage pipeline decouples the learning process: conservative tracing recovers existing stroke geometry under complex degradation, while stroke completion infers reasonable missing strokes.
We compare single-stage and two-stage SFT on the same 60K synthetic samples, with 10,000 steps for each setting.
The single-stage setting directly predicts a complete SVG from the degraded image.
The two-stage setting uses model-predicted O2N outputs as inputs to N2F during evaluation.
As shown in~\cref{tab:ablation_stages}, two-stage SFT improves SSIM from 0.8916 to 0.9055, reduces LPIPS from 0.1210 to 0.1120, and lowers MSE from 0.04153 to 0.03881.
These consistent improvements indicates our two-stage pipeline design is effective under the evaluated training setting. Additional visual comparisons are provided in the supplementary material 2.

\begin{figure}[!t]
	\centering
	\includegraphics[width=\linewidth]{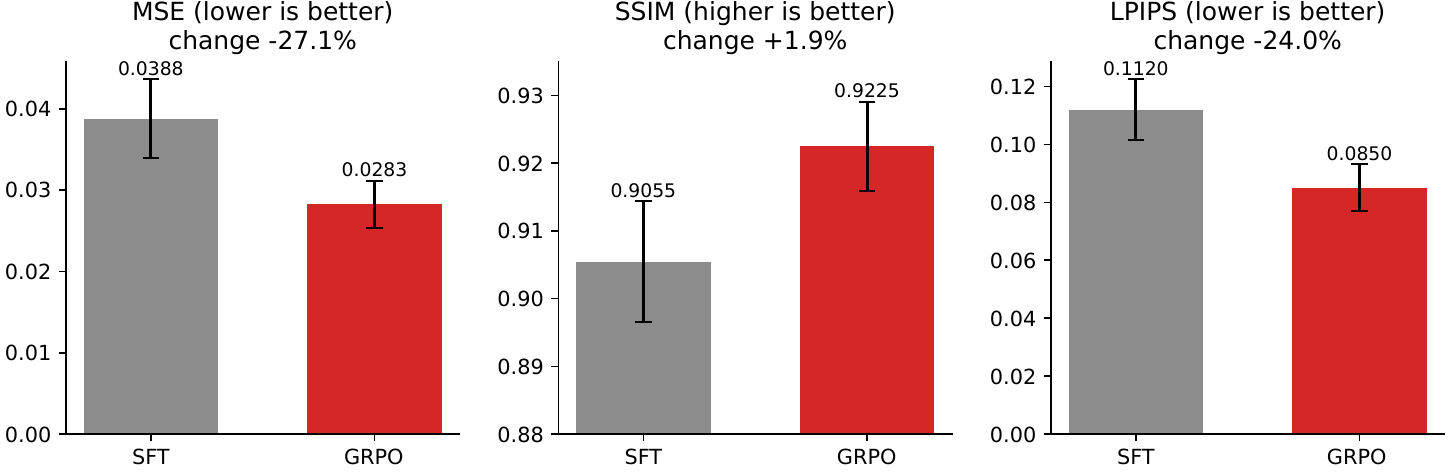}
	\caption{\textbf{Quantitative comparison of SFT and GRPO.} Error bars indicate the standard error of the mean (SEM).}
	\label{fig:rl_metrics}
\end{figure}

\begin{table}[!t]
	\centering
	\caption{Ablation of stage decomposition for character restoration. Single-stage redrawing directly generates the complete SVG; two-stage redrawing uses the full O2N--N2F pipeline.}
	\label{tab:ablation_stages}
	\footnotesize
	\setlength{\tabcolsep}{4pt}
	\begin{tabular*}{\columnwidth}{@{\extracolsep{\fill}}lccc@{}}
		\toprule
		Pipeline Setting & SSIM$\uparrow$ & LPIPS$\downarrow$ & MSE$\downarrow$ \\
		\midrule
		Single-stage complement &0.8916 & 0.1210 &0.04153 \\
		Two-stage complement &0.9055  & 0.1120  & 0.03881 \\
		\bottomrule
	\end{tabular*}
	\vspace{-6pt}
\end{table}

\begin{figure}[!t]
	\centering
	\includegraphics[width=0.9\linewidth]{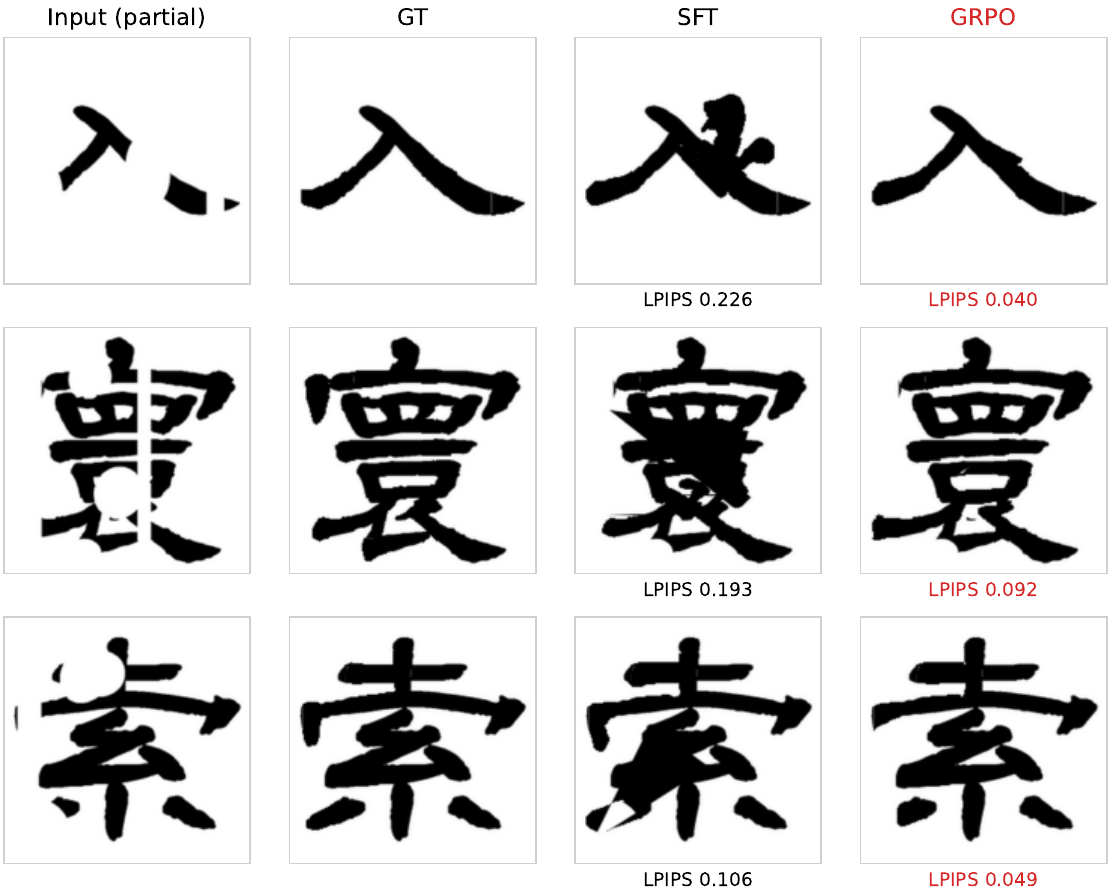}
	\caption{\textbf{Qualitative comparison of SFT and GRPO.} Both variants receive model-predicted O2N partial SVGs as input. The model trained by GRPO can output cleaner and more compliant SVG code for stroke complement.}
	\label{fig:rl_qualitative}
\end{figure}

\textbf{Effectiveness of skeleton-guided reinforcement learning.}
Using the SFT model trained by 60K synthetic data, we evaluate 1,000 additional GRPO updates on N2F task. Note that both variants retain skeleton CoT.
As shown in~\cref{fig:rl_metrics}, GRPO reduces MSE and LPIPS by approximately 27\% and 24\%, respectively, and increases SSIM from 0.9055 to 0.9225.
The examples in~\cref{fig:rl_qualitative} show fewer incorrect connections and redundant additions overlapping existing strokes after GRPO refinement.
The model after GRPO has greatly alleviated these issues. We believe these results are consistent with the goals of our multiple structure-aware rewards and overlap penalty.

\section{Limitations and Future Work}
\label{sec:limitations}
ReDraw has several limitations. First, context windows restrict output sequences at 8k tokens, limiting support for highly complex character. Second, our restoration remains unreliable when more than 50\% of strokes are missing or ink degradation is more diverse (see Supplementary Material 3 for failure cases). Third, ReDraw restores only foreground characters, excluding backgrounds that contain valuable paleographic evidence.
Future work will integrate ReDraw into a full-page document restoration agent, using contextual evidence and an agentic verifier for iterative refinement. Since generated paths are candidate reconstructions rather than verified historical forms, we will explore multiple completion hypotheses with interactive expert assessment to support scholarly use.

\section{Conclusion}
\label{sec:conclusion}
 We present ReDraw, a two-stage VLM for restoring ancient Chinese characters through SVG generation and editing.
 To meet diverse restoration requirements, ReDraw integrates vectorization and tracing into the first stage while placing stroke completion in the second stage. Such a framework alleviates the issues that the previous image restoration model was prone to altering the observed strokes and leading morphological inconsistencies. We also design multiple structure-aware reward functions to achieve GRPO refinement, improving the issue of redundant strokes. A multi-layer synthesis pipeline is developed for Hanjian characters, with the potential to extend to other ancient Chinese character. 
 Extensive experimental results demonstrate ReDraw's outstanding performance in both synthetic and real Hanjian character.
 We anticipate this work will represents a significant advancement in ancient Chinese character restoration and contributes substantially to computational paleography and digital archaeology.

{
    \small
    \bibliographystyle{ieeenat_fullname}
    \bibliography{main}
}

\end{document}